\documentclass[review]{elsarticle}

\usepackage{lineno,hyperref}
\usepackage{marginnote}

\usepackage[printwatermark]{xwatermark}
\usepackage{xcolor}
\usepackage{graphicx}

\begin{document}

\begin{frontmatter}

\title{An Entropic Associative Memory}

\author{Luis A. Pineda\fnref{lapc-footnote}}
\address{Universidad Nacional Aut\'onoma de M\'exico}
\fntext[lapc-footnote]{lpineda@unam.mx}

\author{Gibr\'an Fuentes\fnref{gf-footnote}}
\address{Universidad Nacional Aut\'onoma de M\'exico}
\fntext[gf-footnote]{gibranfp@unam.mx}

\author{Rafael Morales\fnref{rm-footnote}}
\address{Universidad de Guadalajara}
\fntext[rm-footnote]{rmorales@suv.udg.mx}




\begin{abstract}
Natural memories are associative, declarative and distributed. Symbolic computing memories resemble natural memories in their declarative character, and information can be stored and recovered explicitly; however, they lack the associative and distributed properties of natural memories. Sub-symbolic memories developed within the connectionist or artificial neural networks paradigm are associative and distributed, but are unable to express symbolic structure and information cannot be stored and retrieved explicitly; hence, they lack the declarative property. To address this dilemma, we use Relational-Indeterminate Computing to model associative memory registers that hold distributed representations of individual objects. This mode of computing has an intrinsic computing entropy which measures the indeterminacy of representations. This parameter determines the operational characteristics of the memory. Associative registers are embedded in an architecture that maps concrete images expressed in modality-specific buffers into abstract representations, and vice versa, and the memory system as a whole fulfills the three properties of natural memories. The system has been used to model a visual memory holding the representations of hand-written digits, and recognition and recall experiments show that there is a range of entropy values, not too low and not too high, in which associative memory registers have a satisfactory performance. The similarity between the cue and the object recovered in memory retrieve operations depends on the entropy of the memory register holding the representation of the corresponding object. The experiments were implemented in a simulation using a standard computer, but a parallel architecture may be built where the memory operations would take a very reduced number of computing steps.
\end{abstract}


\end{frontmatter}

\newtheorem{mydef}{Definition}

\section{Associative Memory}
\label{sec:memory}

Natural memories of humans and other animals with a developed enough neural system are associative \cite{anderson-1980}. An image, a word or an odor can start a chain of remembrances on the basis of their meanings or contents. Natural memories contrast strongly with standard computer memories in that the latter consists of place-holders --containing strings of symbols that are interpreted as representations-- that are accessed through their addresses. Computational models of associative memories have been extremely difficult to create within the symbolic paradigm, and although there have been important attempts using semantic networks since very early \cite{quillian-1968}, and production systems more recently \cite{Anderson-2004}, practical symbolic associative memories are still lacking. 

This limitation was one of the original motivations for the parallel distributed processing program, including connectionist systems and neural networks \cite{Rumelhart}, which questioned explicitly the capability of Turing Machines to properly address associative memories, among other high level cognitive functions.\footnote{See the introduction of the cited Rumelhart's book.} The subject has been one main subject matter within artificial neural networks and there have been very influential proposals, such as Hopfield's model \cite{hopfield-1982} or the Bidirectional Associative Memory~\cite{bam}. There have been also attempts to represent semantic networks through neural networks \cite{ma-Isahara-2000}. However, neural networks cannot hold symbolic or structured information, and associative memories in this paradigm are rather transfer functions mapping inputs into outputs for classification and prediction among other similar tasks. In a recent model presented by Graves et al. \cite{DBLP:journals/corr/GravesWD14,graves-wayne-nature} the output of the neural network is a vector that is read and written on external tables that can be accessed by location and content, and some symbolic procedures can be learned, but information cannot be stored and/or recovered from a key or a cue, and the contention that such systems are not proper declarative memories still holds \cite{Fodor-Pylyshyn}.  

In this paper we address such limitation and present an associative memory mechanism constituted by a number of associative memory registers that hold distributed representations of objects, and yet these can be registered, recognized and retrieveed on the basis of a cue, and the recovered objects can be expressed declaratively and interpreted as symbolic information.

\section{Relational-Indeterminate Computing}
\label{sec:computing}

The present associative memory systems is defined with a novel mode of computing that is referred to here as Relational-Indeterminate Computing (RIC) \cite{Pineda-ECR-2020,Pineda-modo-comp}. The basic object of computing in this mode is the mathematical relation, such that an object in the domain may be related to several objects in the codomain. The specification is presented by Pineda \cite{Pineda-ECR-2020} as follows:\footnote{For a more general discussion see \cite{Pineda-RC-libro-2020}.}

Let the sets $A = \{a_1,...,a_n\}$ and $V = \{v_1,...,n_m\}$, of cardinalities $n$ and $m$, be the domain and the codomain of a finite relation $r: A\to V$. The objects in the domain and codomain are referred to here as the \emph{arguments} and the \emph{values} respectively. For purposes of notation, for any relation $r$ we define a function $R: A\times V\to \{0,1\}$ --the relation in lower case and the function in upper case letters-- such that $R(a_i,v_j) = 1$ or \emph{true} if the argument $a_i$ is related to the value $v_j$ in $r$, and $R(a_i,v_j) = 0$ or \emph{false} otherwise.

In this formalism, \emph{evaluating a relation} is construed as selecting randomly one among the values associated to the given argument. In the same way that ``$f(a_i) = v_j$'' is interpreted as stating that the value of the function $f$ for the argument $a_i$ is $v_j$, ``$r(a_i) = v_j$'' states that the value of the relation $r$ for the argument $a_i$ is an object $v_j$ that is selected randomly -with an appropriate distribution-- among the values for which $R(a_i,v_j)$ is true.

RIC has three basic operations: \emph{abstraction}, \emph{containment} and \emph{reduction}. Let $r_f$ and $r_a$ be two arbitrary relations from $A$ to $V$, and $f_a$ be a function with the same domain and codomain. The operations are defined as follows:

\begin{itemize}
\item Abstraction: $\lambda(r_f, r_a) = q$, such that $Q(a_i, v_j) = R_f(a_i, v_j) \lor R_a(a_i,v_j)$ for all $a_i \in A$ and $v_j \in V$ --i.e., $\lambda(r_f, r_a) = r_f \cup r_a$.
\item Containment: $\eta(r_a, r_f)$ is true if $R_a(a_i,v_j) \to R_f(a_i,v_j)$ for all $a_i \in A$ and $v_j \in V$ (i.e., material implication), and false otherwise.
\item Reduction: $\beta(f_a, r_f) = f_v$ such that, if $\eta(f_a,r_f)$ holds $f_v(a_i) = r_f(a_i)$ for all $a_i$, where the random distribution is centered around $f_a$, as elaborated below. If $\eta(f_a,r_f)$ does not hold, $\beta(f_a, r_f)$ is undefined --i.e., $f_v(a_i)$ is undefined-- for all $a_i$.
\end{itemize}

Abstraction is a construction operation that produces the union of two relations. A function is a relation and can be an input to the abstraction operation. Any relation can be constructed out of the incremental abstraction of an appropriate  set of functions. The construction can be pictured graphically by overlapping the graphical representation of the included functions on an empty table, such that the columns correspond to the arguments, the rows to the values and the functional relation is depicted by a mark in the intersecting cells. 

The containment operation verifies whether all the values associated to an argument $a_i$ in $r_a$ are associated to the same argument in $r_f$ for all the arguments, such that $r_a \subseteq r_f$. The containment relation is false only in case $R_a(a_i,v_j)=1$ and $R_f(a_i,v_j)=0$ --or if $R_a(a_i,v_j) > R_f(a_i,v_j)$-- for at least one $(a_i,v_j)$. 

The set of functions that are contained in a relation, which is referred to here as the \emph{constituent functions}, may be larger than the set used in its construction. The constituent functions are the combinations that can be formed by taking one value among the ones that the relation assigns to an argument, for all the arguments. The table format allows to perform the abstraction operation by direct manipulation and the containment test by inspection. The construction consists on forming a function by taking a value corresponding to a marked cell of each column, for all values and for all columns. The containment is carried on by verifying whether the table representing the function is contained within the table representing the relation by testing all the corresponding cells through material implication.

For this, the abstraction operation and the containment test are productive. This is analogous to the generalization power of standard supervised machine learning algorithms that recognize not only the objects included in the training set but also other objects that are similar enough to the objects in such set.

Reduction is the functional application operation. If the argument function $f_a$ is contained in the relation $r_f$, reduction generates a new function such that its value for each of its arguments is selected from the values in the relation $r_f$ for the same argument. In the basic case, the selection function is the identity function --i.e., $\beta(f_a, r_f) = f_a$. However, $\beta$ is a constructive operation such that the argument function $f_a$ is the cue for another function recovered from $r_f$, such that $v_j$ is selected from $\{v_j| (a_i, v_j)\in r_f\}$ using and appropriate random distribution function centered around $f_a(a_i)$. If $f_a$ is not contained in $r_f$ the value of such functional application operation is not defined.

Relations have an associated entropy, which is defined here as the average indeterminacy of the relation $r$. Let $\mu_i$ be the number of values assigned to the argument $a_i$ in $r$; let $\nu_i$ = $1/\mu_i$ and $n$ the number of arguments in the domain. In case $r$ is partial, we define $\nu_i = 1$ for all $a_i$ not included in $r$. The \emph{computational entropy} $e(r)$ --or the entropy of a relation-- is defined here as: $$e(r) = -\frac{1}{n} \sum_{i=1}^{n} \log_2(\nu_i).$$

A function is a relation that has at most one value for any of its arguments, and its entropy is zero. Partial functions do not have a value for all the arguments, but this is fully determined and the entropy of partial functions is also zero.

\section{Table Computing}
\label{sec:tables}

The implementation of RIC in table format is referred to as \emph{Table Computing} \cite{Pineda-modo-comp}. The representation consists of a set of tables with $n$ columns and $m$ rows, where each table is an \emph{Associative Register} that contains a relation between a set of arguments $A = \{a_1,..., a_n\}$ and a set of values $V = \{v_1,...,v_m\}$. Let $[R_k]^t$ be the content of the register $R_k$ at time $t$ and $\gets$ an assignment operator such that $R_k \gets R_j$ assigns $[R_j]^t$ to $[R_k]^{t+1}$, where $j$ and $k$ may be equal. This corresponds to the standard assignment operator of imperative programming languages. The machine also includes the conditional operator \emph{if} relating a condition \emph{pred} to the operations $op_1$ and $op_2$  --i.e., if \emph{pred} then $op_1$ else $op_2$, where $op_2$ is optional. The initialization of a register $R$ such that all its cells are set to $0$ or to $1$ is denoted $R\gets0$ or $R\gets1$ respectively, and $R\gets f$ denotes that a function $f:A\to V$ is input into the register $R$. The system also includes the operators $\lambda$, $\eta$ and $\beta$ for computing the corresponding operations. These are all the operations in table computing.

Let $K$ be a class, $O_k$ a set of objects of class $K$, and $F_O$ a set of functions with $n$ arguments and $m$ values, such that each function $f_i \in F_O$ represents a concrete instance $o_i \in O_k$ in terms of $n$ features, each associated to one of $m$ possible discrete values. The function $f_i$ may be partial --i.e., some features may have no value.

Let $R_k$ be an associative memory register and $R_{k-{i/o}}$ an auxiliary input and output register, both of size $n \times m$. The distributed representation of the objects $O_k$ is created by the algorithm $\textit{Memory\_Register}(f_i,  R_k)$ for all $f_i \in F_O$ as follows:  

\begin{itemize}
\item $\textit{Memory\_Register}(f_i,  R_k)$:
\begin{enumerate}
\item $R_{k-{i/o}}\gets f_i$
\item $R_k \gets \lambda(R_k, R_{k-{i/o}})$
\item $R_{k-{i/o}}\gets 0$
\end{enumerate}
\end{itemize}
 
The recognition of an object $o \in O_k$ --or of class $K$-- represented by the function $f$, is performed by the algorithm \emph{Memory\_Recognize} as follows:

\begin{itemize}
\item $\textit{Memory\_Recognize}(f, R_k)$:
\begin{enumerate}
\item $R_{k-{i/o}}\gets f$
\item If $\eta(R_{k-{i/o}},R_k)$ then $(R_{k-{i/o}}\gets1)$
\end{enumerate}
\end{itemize}

The retrieve or recovery of an object is performed through the reduction operation. This is a constructive operation that recovers the content of the memory in the basis of the cue. The procedure is as follows:

\begin{itemize}
\item $\textit{Memory\_Retrieve}(f, R_k)$:
\begin{enumerate}
\item $R_{k-{i/o}}\gets f$
\item If $\eta(R_{k-{i/o}},R_k)$ then $R_{k-{i/o}}\gets \beta(R_{k-{i/o}},R_k)$ else $R_{k-{i/o}}\gets 0$
\end{enumerate}
\end{itemize}

The standard input configuration of this machine specifies that the auxiliary register contains the function to be registered or recognized, or the cue of a retrieve, at the initial state of the corresponding memory operation. This condition is enforced at step (1) of the three memory procedures. The standard output specifies the content of the auxiliary register when the computation is ended. It is $0$ when the memory register operation is ended and when the cue is rejected in a retrieve operation, and $1$ if recognition is successful; otherwise the object to be recognized is rejected and if the retrieve operation is successful it contains the function recovered out of the cue.

The interpretation conventions state that the content of the associative register is interpreted as an abstract concept, and the content of the input auxiliary register is interpreted as a concrete concept --the concept of an individual object.

\section{Architecture}
\label{sec:architecture}

Table Computing was used to implement a novel associative memory system. The conceptual architecture is illustrated in Figure \ref{fig:arq-mem-asoc}. The main components are:

\begin{figure} 
\includegraphics[width=.8\textwidth]{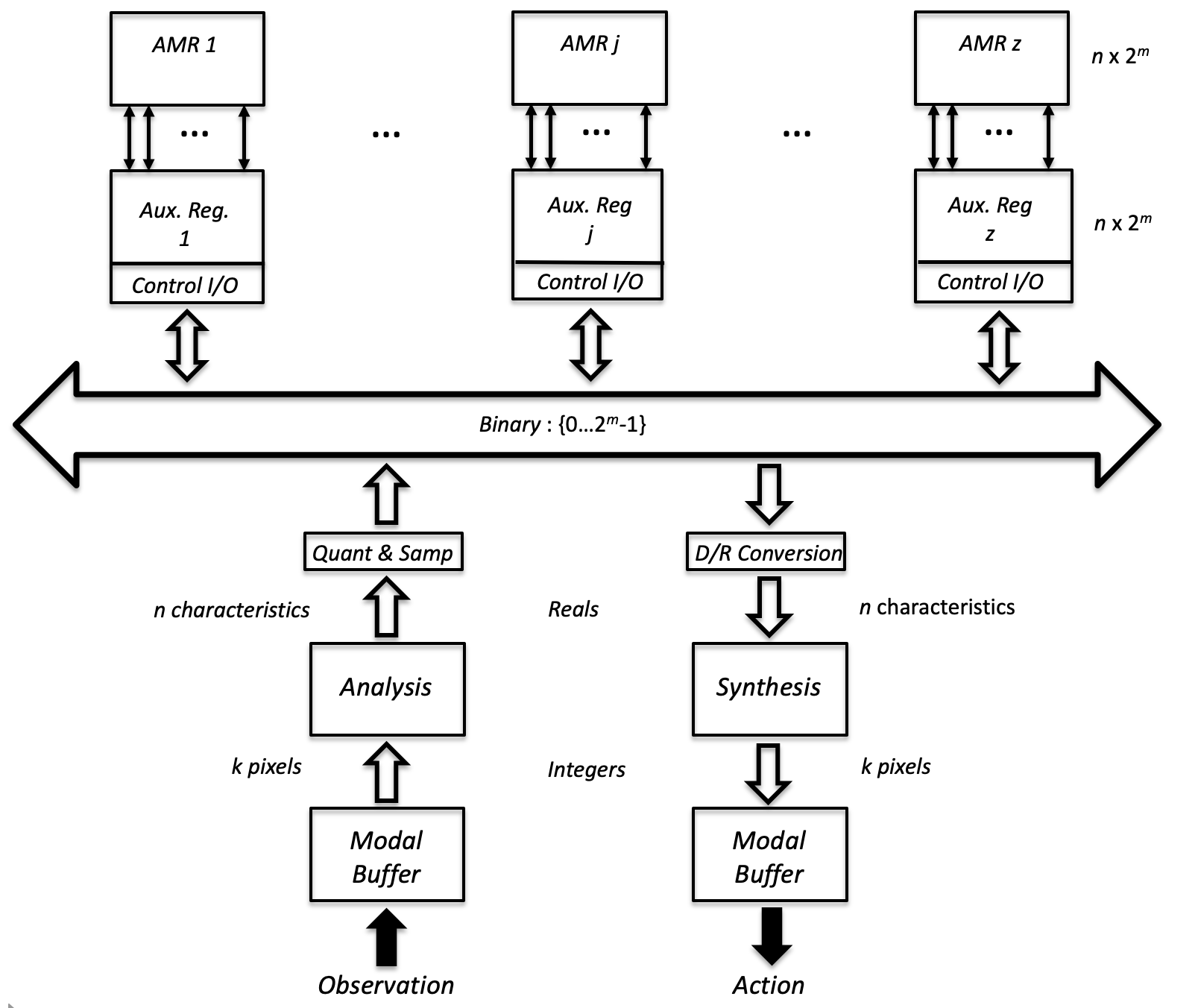}
\centering
\caption{Associative Memory Architecture}
\label{fig:arq-mem-asoc}
\end{figure}

\begin{itemize}
\item  A set of \emph{Associative Memory Registers} (AMRs) of size $n \times 2^m$ for $n \geq 1$ and $m \geq0$ with their corresponding \emph{Auxiliary Registers} (Aux. Reg.) each having a \emph{Control I/O} unit.
\item A central control unit sending the operation to be performed to all AMRs, and receiving the final status of the operation (i.e., whether it was successful or not) and the entropy of each AMR (not shown in the diagram).
\item A bus with $n$ tracks, each representing a characteristic or feature, with its corresponding value: a binary number from $0$ to $2^m-1$ at a particular time.
\item An input processing unit constituted by:
\begin{itemize}
\item A input modal pixel buffer with the concrete representation of images produced by the observations made by the computing agent directly. For instance, the input buffer can contain $k$ pixels with $256$ gray levels represented by integers from $0$ to $255$.
\item An analysis module mapping concrete representations to abstract modality-independent representations constituted by $n$ characteristics with their corresponding real values.
\item A quantizing and sampling module mapping the real values of the $n$ characteristics into $2^m$ levels, represented by binary digits of length $m$, which are written on the corresponding track of the bus.
\end{itemize}
\item An output processing unit constituted by:
\begin{itemize}
\item A Digital/Real conversion module that maps binary numbers in a track of the bus to real numbers, for the $n$ tracks.
\item A synthesis module mapping abstract modality-independent representations constituted by $n$ characteristics with their corresponding values into concrete representations with $k$ pixels with their values.
\item An output modal buffer with the concrete representation of images produced by the synthesis module. The contents of the output buffer are rendered by an appropriate device, and constitute the \emph{actions} of the system.
\end{itemize}

\end{itemize}

The bus holds the representations of functions with domain $A = \{a_1,...,a_n\}$ and range $V = \{v_1,...,v_n\}$ where $0\leq v_j \leq2^m-1$ that represent the objects that are stored, recognized or recovered from the AMRs by the corresponding operations. Objects are placed on the bus through an input protocol; and recovered objects from the AMR are rendered through an output protocol, as follows:

\begin{itemize}
\item \emph{Input Protocol}:

\begin{enumerate}
\item Sense the object $o_i$ and place it in the input modal buffer;
\item Produce its abstract representation $f_i$ through the analysis module;
\item Produce its corresponding quantized representation and write it on the bus;
\item Write the value of all $n$ arguments diagrammatically into the corresponding row of the auxiliary register $R_{k-{i/o}}$;
\end{enumerate}

\item \emph{Output Protocol}:
\begin{enumerate}
\item Write the content of the auxiliary register $R_{k-{i/o}}$ on the bus as binary numbers with $m$ digits for all $n$ arguments;
\item Transform the digital values of the $n$ characteristics into real values;
\item Generate the concrete representation of the object and place it on the output buffer through the synthesis module.
\end{enumerate}

\end{itemize}

The core operations of the memory register, memory recognition and memory retrieve algorithms are carried on directly on the AMRs and their corresponding auxiliary registers in two or three computing steps --i.e., the operations $\lambda$, $\eta$ and $\beta$ in addition to the corresponding assignment operations. The assignments $R_{k-{i/o}}\gets f_i$, $R_{k-{i/o}}\gets 0$ and $R_{k-{i/o}}\gets 1$ are carried out in a single computing step.

The memory operations use the input and output protocols, and are performed as follows:

\begin{itemize}
\item $Memory\_Register(f_i,  R_k)$: The register $AMR_k$ is set on, and the remaining AMRs are set off; the input protocol is performed; the $Memory\_Register$ operation is performed.
\item $Memory\_Recognize(f_i, R_k)$: All AMRs are set on; the input protocol is performed; the $Memory\_Recognize$ operation is performed; all AMRs send its status and entropy to the control unit; if no AMR's recognition operation is successful the object is rejected.
\item $Memory\_Retrieve(f_i, R_k)$: All AMRs are set on; the input protocol is performed; the $Memory\_Retrieve$ operation is performed; all AMRs send its status and entropy to the control unit; all AMRs but the selected one are set off; the output protocol is executed; the recovered object is placed on the output buffer.

\end{itemize}

\section{Analysis and Synthesis}
\label{sec:analisis-sintesis}

The Analysis module maps the concrete information that is sensed from the environment and placed in the input buffer --where the characteristics stand for external signals-- into the abstract representation that characterizes the membership of the object within a class. Both concrete and abstract representations are expressed as functions but while in the former case the arguments have a spatial interpretation --such as the pixels of an image-- in the latter the functions stand for modality-independent information.

\begin{figure} 
\includegraphics[width=0.8\textwidth]{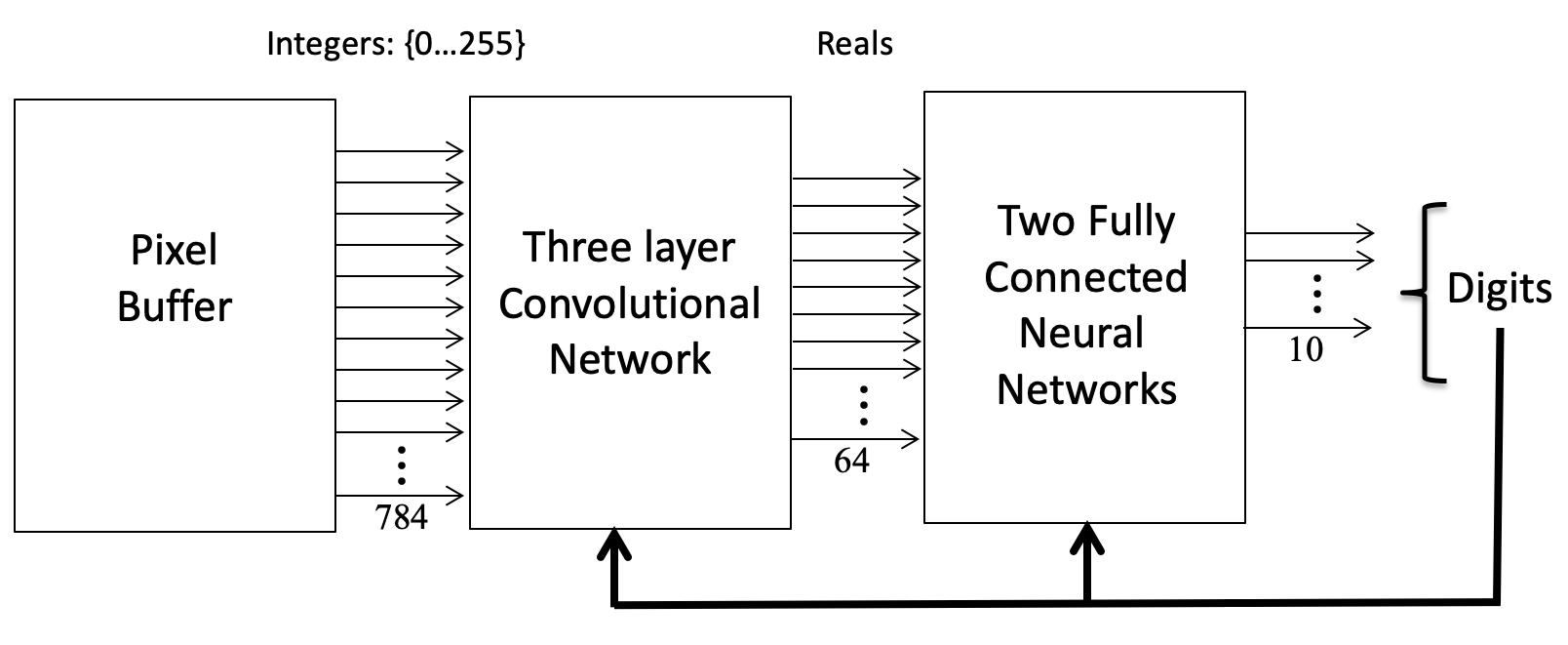}
\centering
\caption{Training the Analysis Module}
\label{fig:training-architecture}
\end{figure}

The analysis module in the present architecture is constituted by a neural network with three convolutional layers \cite{Lecun-nature}. The training phase was configured by adding a classifier, which was a fully connected neural network (FCNN) with two layers. The analysis module was trained in a standard supervised manner with back-propagation, as illustrated in Figure \ref{fig:training-architecture}.

Once the analysis module has been trained the FCNN is removed and the objects in the input buffer can be mapped into their corresponding representation as a set of abstract features through a bottom-up ``analysis operation''. The information is feed into the AMRs through the output of the convolutional layer. The purpose of the analysis module is to map concrete images into abstract representations, but not to perform classification.

The diagram shows the case in which the input has $784$ inputs, corresponding to a pixel buffer of size $28 \times 28$, each taking one out of $256$ gray levels, while its output is a function with $64$ arguments with real values.
 
The objects recovered from the AMRs are mapped into the corresponding concrete representations and placed in the output buffer by the synthesis module. This consists of a transposed convolutional network that computes the inverse function of the input convolutional network. The two neural networks together conform what is known as a convolutional autoencoder \cite{hintonReducingDimensionalityData2006,masciStackedConvolutionalAutoEncoders2011}.

If the function produced by the analysis module is input directly into the synthesis module, the image in the output buffer should be the same as the one originally placed in the input buffer. However, the transposed convulational network was trained independently and the recovered image is slightly different.

\section{A Visual Memory for Hand Written Digits}
\label{sec:experiment}

The associative memory system was simulated through the construction of a visual memory for storing distributed representations of hand written digits from ``0'' to ``9''. The system was built and tested using the MNIST database.\footnote{http://yann.lecun.com/exdb/mnist/} In this resource each digit is defined as a 28 x 28 pixel array with 256 gray levels. There are 70,000 instances available. The instances of the ten digit types are mostly balanced.

The corpus was partitioned further in three disjoint sets: 

\begin{itemize}
    \item Training Corpus (\emph{TrainCorpus}): For training the analysis and synthesis convolutional and transposed networks (57 \%).
    \item Remembered Corpus (\emph{RemCorpus}): For filling in the Associative Memory Registers (33 \%).
    \item Test Corpus (\emph{TestCorpus}): For testing (10 \%).
\end{itemize}

The corpus partitions were rotated through a standard 10-fold cross-validation procedure.

Four experiments supported by a given analysis and synthesis modules were performed:

\begin{enumerate}
\item Experiment 1: Define an associative memory system including an AMR for holding the distributed representation of each one of the ten digits. Determine the precision and recall of the individual AMRs and of the overall system. Identify the size of the AMRs with satisfactory performance.
\item Experiment 2: Investigate whether AMRs can hold distributed representations of more than one individual object. For this an associative memory system including an AMR for holding the distributed representation of two ``overlapped'' digits is defined. Determine the precision and recall of the individual AMRs and of the overall system.
\item Experiment 3:  Determine the overall precision and recall for different levels of entropy of the AMRs, for the AMR with the best performance identified in Experiment 1.
\item Experiment 4: Retrieve objects out of a cue for different levels of entropy and generate their corresponding images --with the same AMR used in experiment 3. Assess the similarity between the cue and the recovered object at different levels of entropy. 
\end{enumerate}

In all four experiments each instance digit is mapped into a set of $64$ features through the analysis module. Hence, each instance is represented as a function $f_i$ with domain $\{a_1,...,a_{64}\}$ where each argument $a_i$ is mapped to a real value $v_i$ --i.e. $f_i(a_i) = v_j$. The values are quantized in $2^m$ levels. The tables or associative registers have sizes of $64 \times 2^m$. The parameter $m$ determines the granularity of the table. The experiments one and two were performed with granularities $2^m$ for $0<=m<=9$. So, the memory was tested with 10 granularities in each setting. The source code for replicating the experiments, including the detailed results and the specifications of the hardware used, are available in Github at \url{https://github.com/LA-Pineda/Associative-Memory-Experiments.}

\subsection{Experiment 1}
\label{sec:exp-1}

Compute the characteristics of AMR of size $64 \times 2^m$ for $0\leq m \leq9$:

\begin{enumerate}
    \item Register the totality of \emph{RemCorpus} in their corresponding register through the $Memory\_Register$ operation;
    \item Test the recognition performance of all the instances of the test corpus through the $Memory\_Recognize$ operation;
    \item Compute the average precision, recall and entropy of individual memories.
    \item Select a unique object to be recovered by the $Memory\_Retrieve$ operation; compute the average precision and recall of the integrated system when this choice has been made.
\end{enumerate}
 
\begin{figure} 
\includegraphics[width=0.8\textwidth]{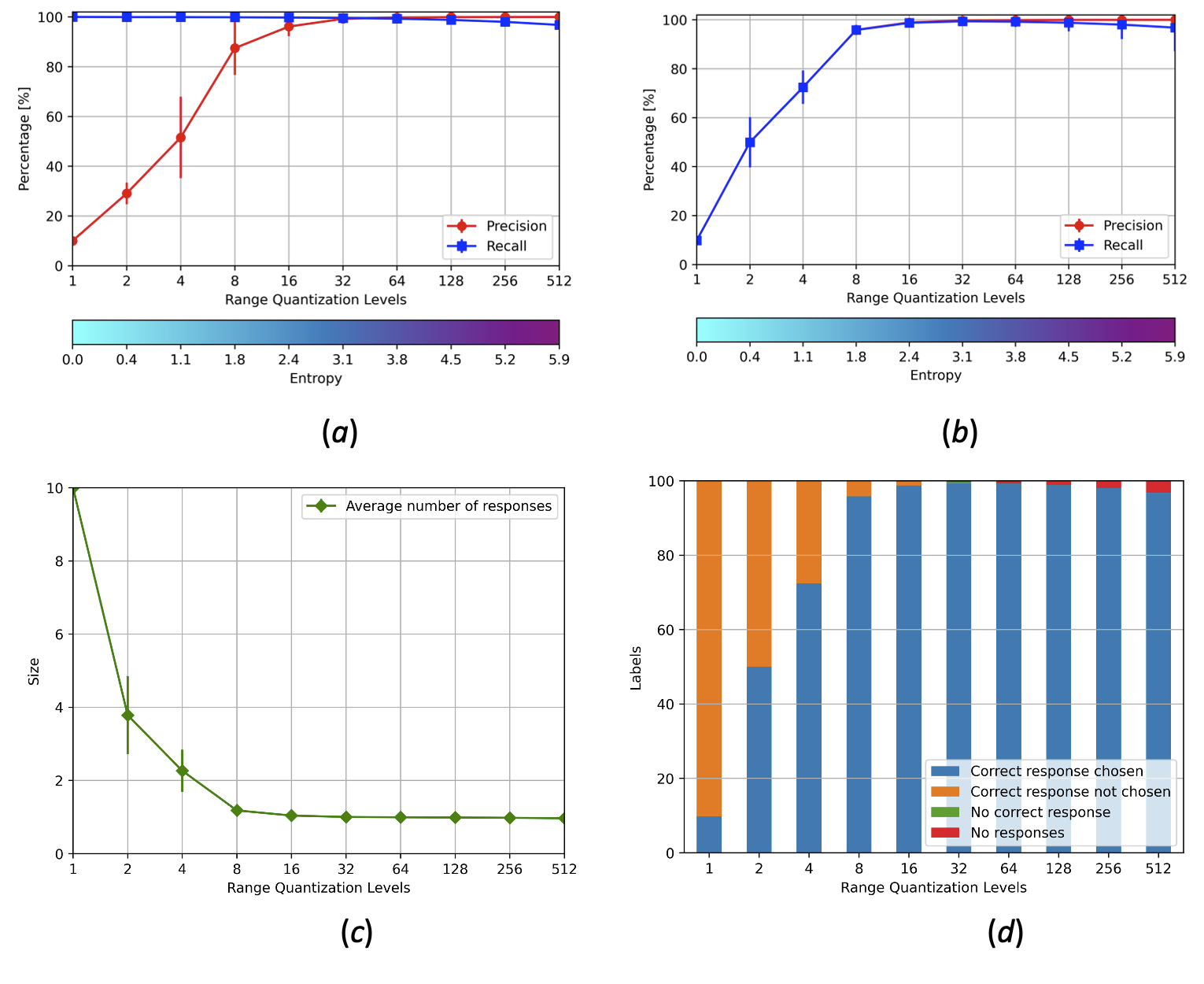}
\centering
\caption{Results of Experiment 1}
\label{fig:experiment-1}
\end{figure}

The average precision, recall and entropy of the ten AMRs are shown in \ref{fig:experiment-1} (a). The precision for the smallest AMR with only one row are $10 \%$ --the proportion of the test data of each digit-- and recall is $100 \%$ --as all the information is confused and everything is accepted. The precision grows with the size of the AMRs and has a very satisfactory value up from $32$ rows. The recall, on its part, remains very high until the granularity of the table is too fine and it starts to decrease slightly. The entropy is increased almost linearly with the AMRs size, starting from $0$ where the relations have only one value.
 
The average precision, recall and entropy of the integrated system is shown in \ref{fig:experiment-1} (b). The precision has a similar pattern to the one above, but the recall lowers significantly in AMRs with a small $m$ --the precision and recall are the same practically for $m \leq 4$. The reason of this decrease is that when the size of the AMR is small, there is a large number of false positives, and several AMRs different from the right one may accept the object; however, one register must be selected for the memory retrieve operation, and there is no information to decide which one. This decision was made using the AMR with the minimal entropy, although this choice does not improve over a random choice using a normal distribution.
 
The average number of accepting AMRs for each instance per AMR size is shown in Figure \ref{fig:experiment-1} (c). As can be seen this number goes from $10$ for AMRs with one row to $1$ for AMRs with 8 and 16 rows, where the precision is very high because every AMR recognizes only one instance in average. This effect is further illustrated in Figure \ref{fig:experiment-1} (d).

\subsection{Experiment 2}
\label{sec:exp-2}

\begin{figure} 
\includegraphics[width=0.8\textwidth]{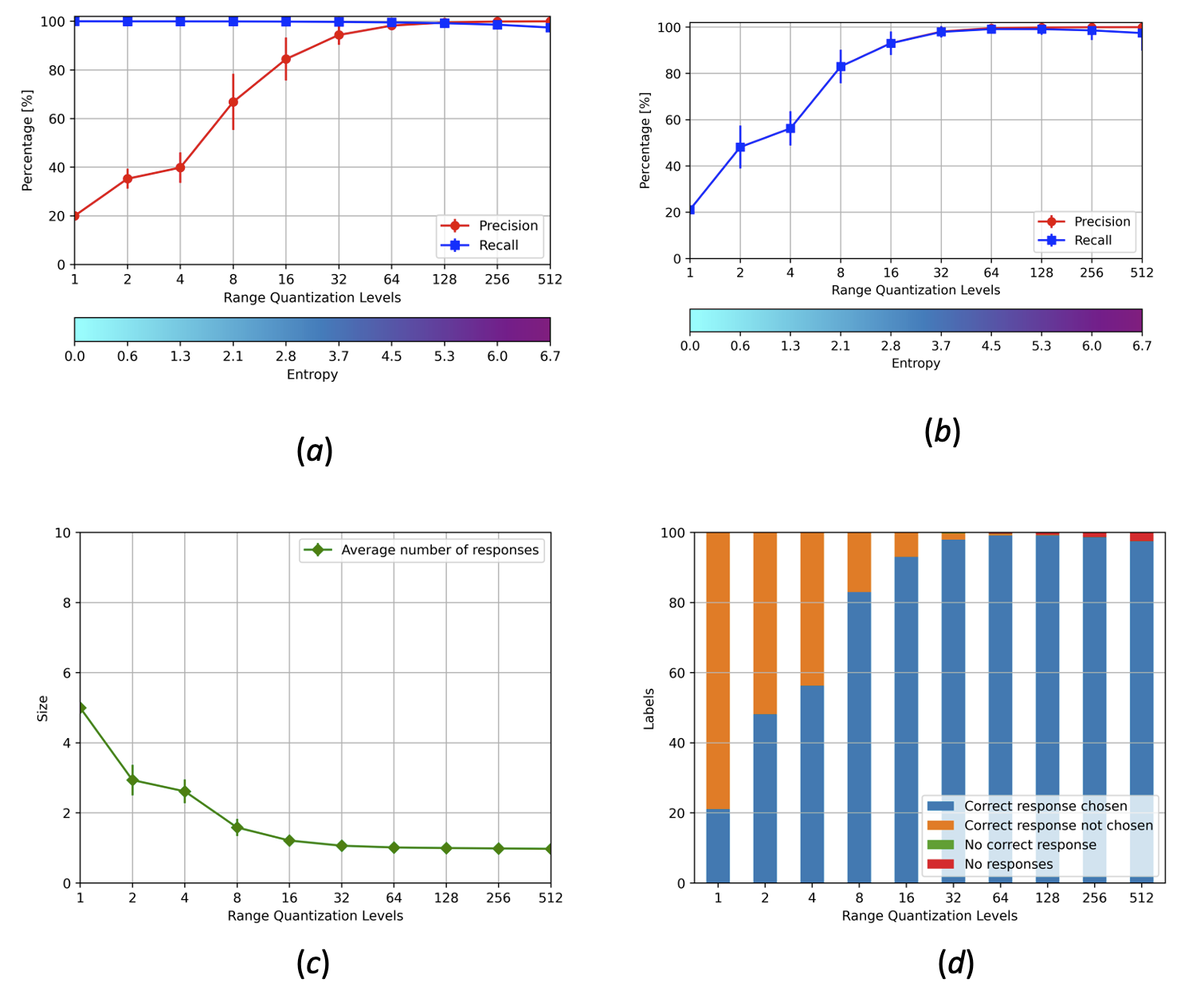}
\centering
\caption{Results of Experiment 2}
\label{fig:experiment-2}
\end{figure}

In this experiment each associative register holds the representation of two different digits:  ``0" and ``1", ``2" and ``3", ``4" and ``5", ``6" and ``7" and ``8" and ``9". The procedure is analogous to experiment 1. The results of the experiment are shown in \ref{fig:experiment-2}. The results of both experiments are also analogous, with the only difference that the entropy of the AMRs holding two digits are larger than the entropies of the AMRs holding a single digit. This experiment shows that it is possible to create associative memories holding overlapped distributed representations of more than one individual object, that have a satisfactory performance.

\subsection{Experiment 3}
\label{sec:exp-3}

The purpose of this experiment was to investigate the performance of an AMR with satisfactory operational characteristics in relation to its entropy or information content. Experiment 1 shows that AMRs with sizes $64 \times 32$ and $64 \times 64$ satisfy this requirement. As their performance are practically the same, the smallest one was chosen for a basic economy criteria.

\begin{figure} 
\includegraphics[width=0.6\textwidth]{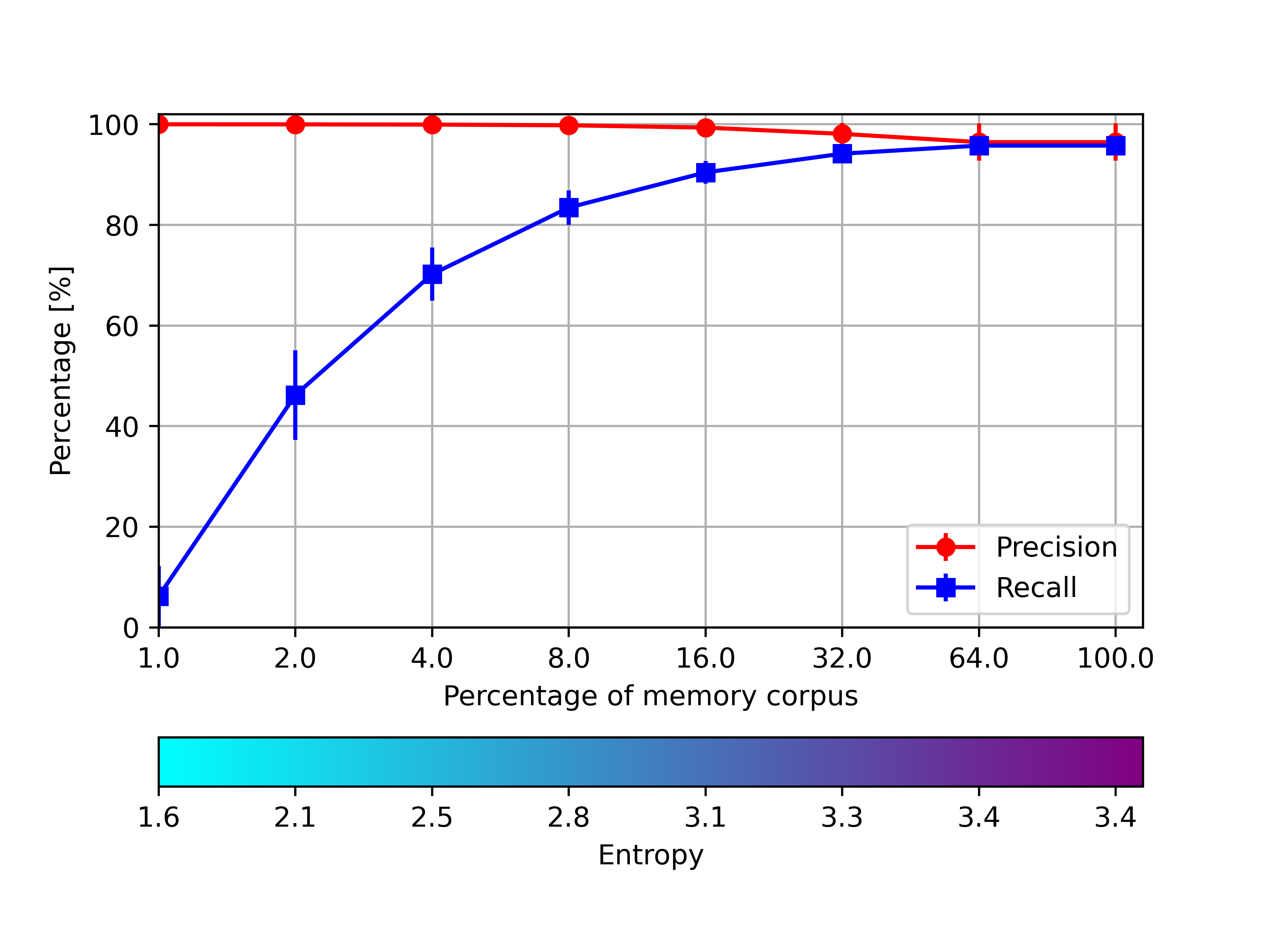}
\centering
\caption{Results of Experiment 3}
\label{fig:experiment-3}
\end{figure}

The AMRs were filled up with varying proportions of the \emph{RemCorpus} --1 \%, 2 \%, 4 \%, 8 \%, 16 \%, 32 \%, 64 \% and 100 \%-- as shown in Figure \ref{fig:experiment-3}. The entropy increases according to the amount of remembered data, as expected. Precision is very high for very low entropy values and it decreases slightly when the entropy is increased, although it remains very high when the whole of \emph{RemCorpus} is considered. Recall, on its part, is very low for very low levels of entropy but increases very rapidly when the AMR is filled up with more data.

\subsection{Experiment 4}
\label{sec:exp-4}

The final experiment consists on assessing the similarity of the objects retrieved from the memory out of a cue. In the basic scenario, if the cue matches perfectly the recovered object, the image in the output buffer should be the same as the image placed in the input. However, memory retrieve is a constructive operation that renders a novel object which may be somewhat different than the cue. The object is constructed by the $\beta$ operation, as described above. In the present experiment a random triangular distribution is used for selecting the values of the arguments of the retrieved object out of the potential values of the AMR for the corresponding arguments.

The hypothesis is that the increase of memory recall goes in hand with higher entropy, but the space of indeterminacy of the AMR impacts negatively in the resemblance or similarity between the cue and the retrieved object. Figure \ref{fig:experiment-3} suggest that this effect is significant for memories with a low entropy --i.e., $e \leq 2.5$.

\begin{figure} 
\includegraphics[width=0.8\textwidth]{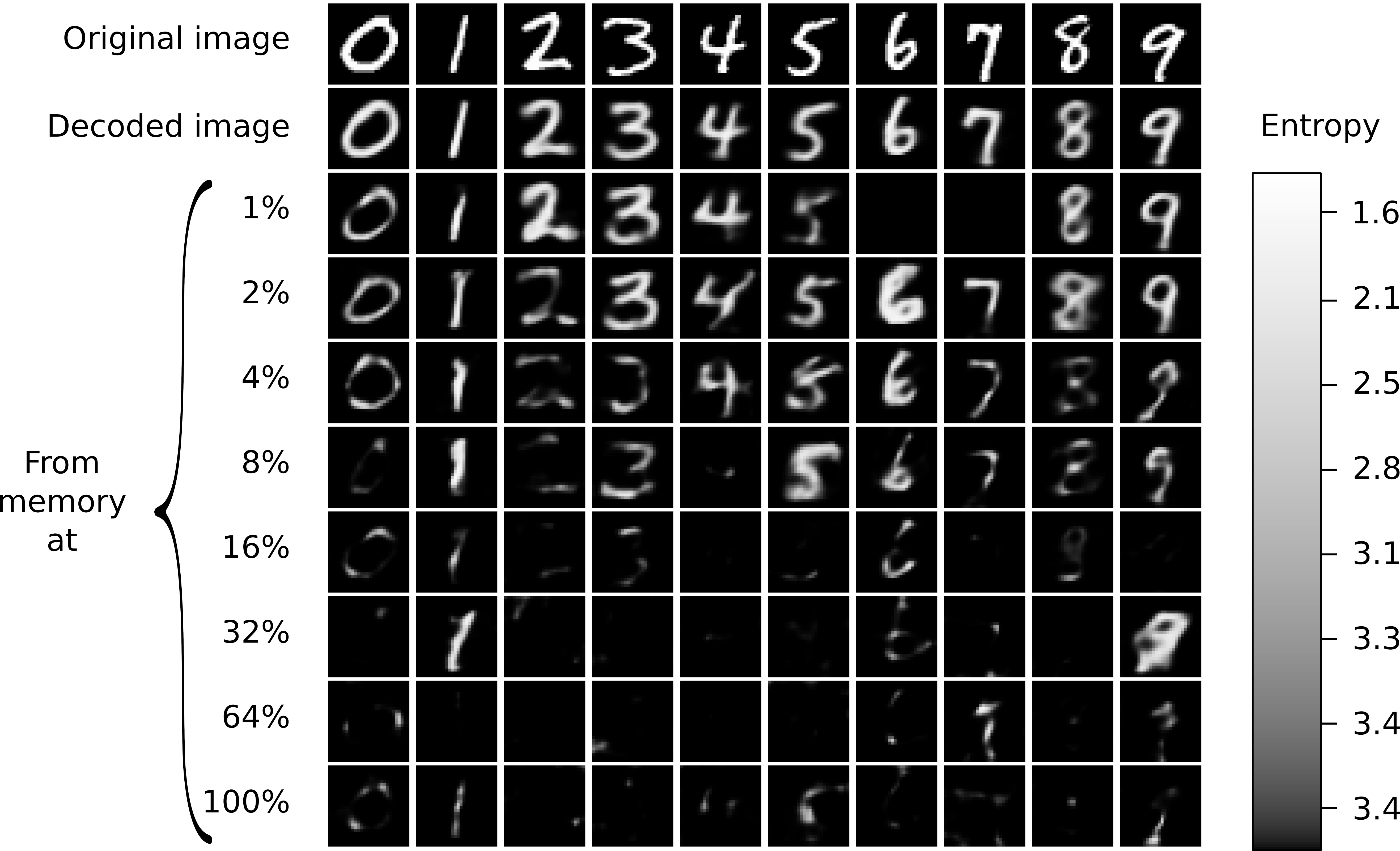}
\centering
\caption{Similarity between the cue and the recovered digits as a function of the entropy}
\label{fig:experiment-4}
\end{figure}

The results of this experiment are shown in Figure \ref{fig:experiment-4}. The first row contains the cue for the retrieval operation for the ten digits; the second is the decoded image, which corresponds to the one resulting from synthesizing the output of the analysis module directly. This is equivalent to specify $\beta$ as the identity function --although such choice would remove the constructive aspect of the memory retrieve operation, and memory recognition and memory retrieve would amount to the same operation. The decoded image is very similar to the cue, but it is not an exact copy. The synthesis module should compute the inverse function of the one computed by the analysis one but the convolutional and the transposed networks are trained independently, and this is only an approximation.

The remaining images, from top to bottom, correspond to the retrieved objects for the nine levels of the \emph{RemCorpus} that are considered (the codified image corresponds to $e = 0$). The rows for the ten digits suggest that the highest similarity is achieved when the entropy is very low. 

The overall behavior of the system suggests that the AMRs are very tolerant for memory recognition, but very restrictive for retrieving objects. This is consistent to the general intuition that recognizing objects in memory is much easier than retrieving them.

\section{Experimental Setting}
\label{sec:setting}
The programming for all the experiments was carried out in Python 3.8 on the Anaconda distribution. The neural networks were implemented with TensorFlow 2.3.0, and most of the graphs produced using Matplotlib. The experiments were run on an Alienware Aurora R5 with an Intel Core i7-6700 Processor, 16 GBytes of RAM and NVIDIA GeForce GTX 1080 graphics card. The images shown in Figure~\ref{fig:experiment-4} were selected one column per experimental run, while the criteria for selection was, when possible, to have some resemblance to the corresponding digit in the first (1\%) and last (100\%) stage.

Regarding the neural networks, the classifier and the decoder were trained separately. Firstly, the classifier was trained (convolutional section plus fully connected section) using MNIST data. Secondly, the fully connected layers were removed from the classifier and the features for all images in MNIST were generated and stored using the trained convolutional part only. Thirdly, the decoder (transposed convolutional) was trained using the features as input data and the original images as labels. Finally, the trained decoder was used to generate the images from the features produced by the memory retrieve algorithm in Experiment~4. 

\section{Discussion}
\label{sec:discussion}

In this paper a memory system that is associative and distributed but declarative is presented. Individual instances of represented objects are characterized at three different levels: i) as concrete modality-specific representations in the input and output buffers --which can be sensed or rendered directly-- or, alternatively, as functions from pixels to values; ii) as abstract modality-independent representations in a space of characteristics, which are functions standing in a one-to-one relation to their corresponding concrete representations in the first level; and iii) as distributed representations holding the disjunctive abstraction of a number, possibly large, of instance objects expressed in the second level. The first level can be considered as declarative and symbolic; the second is still declarative but is independent of representation, so can hold and integrate representations of objects presented in different modalities; and the third is a sub-symbolic structure holding the abstraction of a set of objects of the second level.

The memory register and recognition operations use only the logical disjunction and material implication, that are performed by direct manipulation, cell to cell in the tables, and information is taken and placed on the bus by direct manipulation too, enhancing the declarative aspect of the system.

The associative property depends on the dual role played by the intermediate representations that express content and at the same time select their corresponding Associative Memory Registers through the memory recognition and recovery operations. The memory register operation is analogous to the training phase of supervised machine learning, and it presupposes an attention mechanism that selects the AMR in which the information is input. Adressing this restriction is left for further work.

The analysis and synthesis modules mapping concrete into abstract representations and vice versa are serial computations from a conceptual perspective --although their internal micro-operations can be performed in parallel using GPUs-- but the memory operations manipulate the symbols stored in the corresponding cells of the tables directly, taking very few computing steps, which can be performed in parallel if the appropriate hardware is made available. In the present architecture the memory operations involve the simultaneous activation of all the associative memory registers, and this parallelism takes places not only at the algorithmic and implementation levels but also at the computational or functional system level, in Marr's sense \cite{Marr}.

The analysis and synthesis mechanisms are implemented here through standard deep neural networks, but this is a contingency. From the conceptual perspective this functionality can be achieved with other modes of computing, that map concrete representations into the abstract characteristics space by other means.

The functionality of the memory proper can be also distinguished from the input and output processes in terms of the indeterminacy of the computing objects. The analysis and synthesis modules compute a function whose domain and range are sets of functions, and these processes are fully determined: provide the same value for the same argument always. Hence, these are zero entropy computations. However, the distributed representations stored in the memory registers have a degree of indeterminacy, which is measured by the computing entropy. 

The entropy is a parameter of the system performance, as can be seen in the four experiments. First, it measures the operational range of the associative registers, as shown in experiments 1 and 2. If the entropy is too low precision and recall are low overall, but if it is too high, recall is also diminished. However, there is an entropy level in which both precision and recall are pretty satisfactory. The experiments showed that memory registers with sizes of $64 \times 32$ and $64 \times 64$, with entropy of $3.1$ and $3.8$ respectively, have satisfactory operational characteristics. The smaller register was chosen for the further experiments due to basic economy considerations. 

The experiment 2 shows that a single associative memory register can hold the distributed representation of more than one object. The cost is that the entropy is increased, and larger registers with a large amount of information are required for the construction of operational memories. However, this functionality is essential, both for the construction of higher abstractions, and possibly for the definition of composite concepts. This is also left for further work. The point to stress here is that the measure involved is again the entropy.

The experiment 3 addressed the question of what is the amount of information and the level of entropy that is required for effective memory recognition and retrieval, given an operational memory register. The results show that recognition precision is very high regardless the amount of information that is feed into the memory register. Hence, whenever something is accepted one can be pretty sure that it belongs to the corresponding class. However, recognition recall is very low for low levels of entropy but becomes very high even with a moderate amount of information. Again, for the present domain, the register is useful for entropies of around $3.4$. However, if the information is increased with noise, the entropy has a very large value, and although recognition recall will not decrease, the information is confused and recognition precision will lower significantly. Hence, there is again a range of entropy, not too low and not too high, in which the amount of information is rich, and the memory is effective.

The experiment 4 asked the question of how similar are the objects recovered from the memory retrieval operation to the cue or key used as the retrieval descriptor. The results show that high similarity is only achieved for very low levels of entropy. In the basic case, when the entropy is zero, the retrieved object is the same as the cue, and memory recognition and memory retrieval are not distinguished. This corresponds to \emph{Ramdom Access Memories} (RAM) of standard digital computers, where the content of a RAM register is ``copied'' but not really extracted or recovered in a memory read operation.

Natural memories are constructive in the sense that the memory retrieve operation renders a genuine novel object. This is the reason to define the $\beta$ operator using a random distribution. Whenever the cue is accepted the retrieval operation selects an object whose representation is within the relation's constituent functions. The retrieved object may or may not have been registered explicitly, but it is the product of a construction operation always.

However, the similarity experiment showed that high resemblance between the cue and the recovered object is only achieved when the entropy has very low values. If the entropy is zero, the retrieved object is a ``photographic copy'' --Figure 6 shows some distortions, but these are due to the disparity between the analysis and synthesis modules. The reconstructions resemble well the cue for entropy around $2$ --using only $1 \%$ or $2 \%$ of the test corpus, but then on the similarity is quite random. This result suggests that although there is flexibility in memory recognition, memory retrieval is quite constrained, and hence a much harder operation. The entropy range also suggests that retrieval goes from ``photographic'' to ``recovered objects'' to ``imaged objects'' to noise. Once again, operational memories have an entropy range in which the entropy is not too low and not too high. This pattern seems to be very general and is referred to as \emph{The Entropy Trade-off} \cite{Pineda-ECR-2020}.

The study of memory mechanisms for storing the representations of individual objects is central to cognition and computer applications, such as information systems and robotics. The sense data may be presented to cognition in a natural format in which spatial and temporal information may be accessed directly, but may be stored as a highly abstract modality-independent representation. Such representations may be retrieveed directly by perception in the production of interpretations, by thought in decision making and planning, and by the motricity or motor module when motor abilities are deployed.

Associative memory mechanisms should support long term memory, both episodic and semantic, and may be accessed on demand by working memory. Such devices may be used in the construction of episodic memories and composite concepts that may be stored in associative memory registers themselves, or in higher level structures that rely on basic memory units. Associative memory models may be essential for the construction of lexicons, encyclopedic memories, and modality-specific memories, such as faces, prototypical shapes or voices, both in cognitive studies and applications.

The present investigation addressed the design and construction of the full associative memory system for a simple domain, and the current result can be seen as a proof of concept. We left the investigation of larger and more realistic domains for further work.

The present investigation addressed also the case in which images where complete objects; however, this is only the basic condition, as interpretation is often performed in noisy environments and with incomplete information. For instance, the objects may be partially covered and/or seen from different perspectives. The cues or descriptors generated in such conditions would be much poorer information, memory recognition would be harder, and the objects recovered from memory would involve larger reconstructions than the present case. The investigation of the associative memory system for incomplete information is also left for further research.

\section{Acknowledgments}

We thank Iv\'an Torres and Ra\'ul Peralta for their help in the implementation of preliminary recognition experiments. The first author also acknowledges the partial support of grant PAPIIT-UNAM IN112819, M\'exico.

\nolinenumbers



\bibliography{references}

\end{document}